\documentclass[sigconf,natbib=true]{acmart}

\usepackage{graphicx}
\usepackage{balance}
\usepackage{enumerate}
\usepackage{graphics}
\usepackage{amsmath}
\usepackage{booktabs}
\usepackage{multirow}
\usepackage{xcolor}
\usepackage{enumerate}
\usepackage{makecell}

\usepackage{tabularx}
\usepackage{pifont}
\usepackage{enumitem}
\usepackage{colortbl}
\usepackage{adjustbox}

\AtBeginDocument{%
  \providecommand\BibTeX{{%
    \normalfont B\kern-0.5em{\scshape i\kern-0.25em b}\kern-0.8em\TeX}}}

\copyrightyear{2023}
\acmYear{2023}
\acmConference[Gen-IR@SIGIR2023]{The First Workshop on Generative Information Retrieval}{July 27, 2023}{Taipei, Taiwan}
\acmBooktitle{The First Workshop on Generative Information Retrieval (Gen-IR@SIGIR23), July 27, 2023, Taipei, Taiwan}
\acmDOI{}
\acmPrice{}
\acmISBN{}
\setcopyright{rightsretained}

\newcommand{\dataname}{KI-QFS}

\newcommand{\querysum}{\textsc{QuerySum}}
\newcommand{\margesum}{\textsc{MargeSum}}

\newcommand{\corpus}{\textsc{Corpus}}
\newcommand{\corpusint}{\corpus{}\textsubscript{int}}
\newcommand{\corpusext}{\corpus{}\textsubscript{ext}}
\newcommand{\corpusaug}{\corpus{}\textsubscript{aug}}

\begin{document}

\title{Tackling Query-Focused Summarization as A Knowledge-Intensive Task: A Pilot Study}
\author{Weijia Zhang}
\email{w.zhang2@uva.nl}
\affiliation{%
  \institution{University of Amsterdam}
  \country{Netherlands}
}

\author{Svitlana Vakulenko}
\email{svitlana.vakulenko@gmail.com}
\affiliation{%
  \institution{Amazon Alexa AI}
  \country{Spain}
}

\author{Thilina Rajapakse}
\email{t.c.r.rajapakse@uva.nl}
\affiliation{%
  \institution{University of Amsterdam}
  \country{Netherlands}
}

\author{Yumo Xu}
\email{yumo.xu@ed.ac.uk}
\affiliation{%
  \institution{University of Edinburgh}
  \country{Edinburgh}
}

\author{Evangelos Kanoulas}
\email{E.Kanoulas@uva.nl}
\affiliation{%
  \institution{University of Amsterdam}
  \country{Netherlands}
}

\begin{abstract}

Query-focused summarization (QFS) requires generating a summary given a query using a set of relevant documents.
However, such relevant documents should be annotated manually and thus are not readily available in realistic scenarios.
To address this limitation, we tackle the QFS task as a knowledge-intensive (KI) task without access to any relevant documents. Instead, we assume that these documents are present in a large-scale knowledge corpus and should be retrieved first. To explore this new setting, we build a new dataset (\dataname{}) by adapting existing QFS datasets. In this dataset, answering the query requires document retrieval from a knowledge corpus. We construct three different knowledge corpora, and we further provide relevance annotations to enable retrieval evaluation. Finally, we benchmark the dataset with state-of-the-art QFS models and retrieval-enhanced models. The experimental results demonstrate that QFS models perform significantly worse on \dataname{} compared to the original QFS task, indicating that the knowledge-intensive setting is much more challenging and offers substantial room for improvement. We believe that our investigation will inspire further research into addressing QFS in more realistic scenarios.

\end{abstract}



\maketitle

\begin{figure}[t]
\centering
\includegraphics[width=0.47\textwidth]{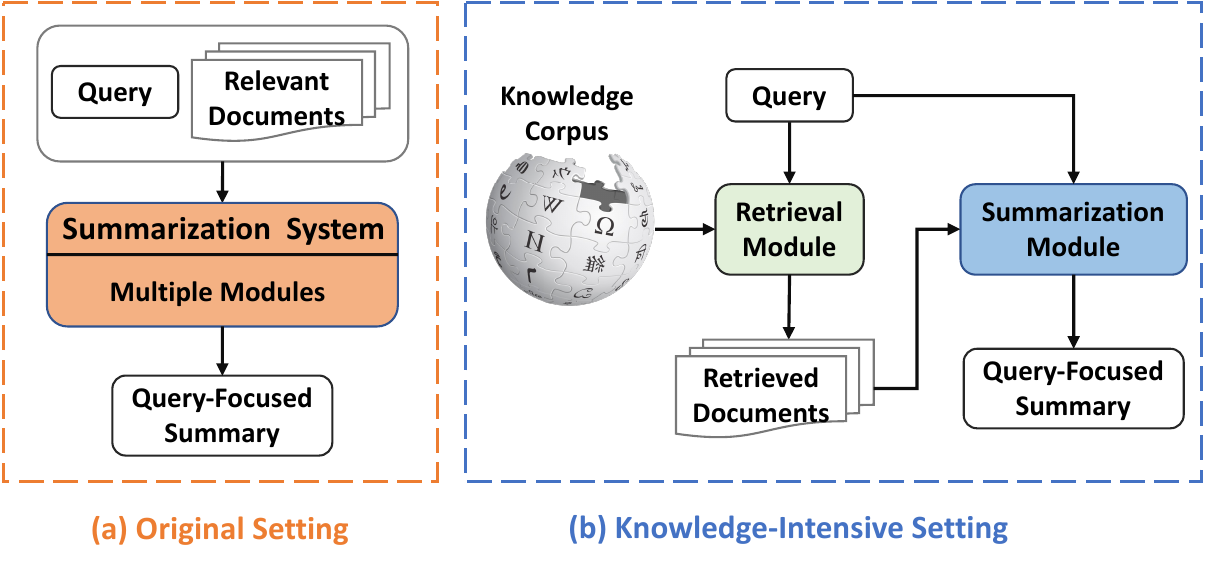}
\caption{The Comparison between (a) the original QFS and (b) our knowledge-intensive setting. The original QFS assumes relevant documents are available. While our setting aims to generate the summary without relying on such relevant documents.}
\label{fig:comparison}
\end{figure}

\section{Introduction}

Query-focused summarization (QFS) \cite{dang2006duc} aims to produce a multi-sentence summary from a set of topic-related documents to answer a given query. 
This task has important applications in various domains such as search engines and report generation \cite{ma2023mdssurvey}. Although recent years have witnessed the advancement of QFS methods \cite{xu2020querysum,laskar2020wsl,xu2020margesum} thanks to large pre-trained language models \cite{devlin2019bert,lewis2020bart},
most existing QFS datasets, such as widely-used DUC 2005-2007 datasets \cite{dang2006duc}, are still relatively small in size and mainly used as evaluation-only sets, due to the labor-intensive nature of creating QFS datasets. 
For instance, in the DUC datasets, human annotators should first collect relevant documents and then carefully write a summary to answer a given query based on the documents. More importantly, the assumption that relevant documents are readily available is often impractical, which hinders the further development of QFS methods.

In this work, we tackle the QFS task as a knowledge-intensive (KI) task \cite{petroni2021kilt}, where relevant documents should be retrieved from a large-scale knowledge corpus given a query. Figure \ref{fig:comparison} shows the difference between the original QFS and our knowledge-intensive setting. In the original setting, relevant documents are available as inputs. Note that the summarization system is often complex and consists of different modules. For instance, it may contain a relevance module that selects salient sentences and a generation module that produces the summary based on the sentences. While our setting addresses QFS in a more realistic scenario, where relevant documents are present in a knowledge corpus. In our setting, we employ an explicit retrieval module to retrieve more concise documents from a large-scale knowledge corpus and feed them into a summarization module to obtain the summary. 
In order to explore this new knowledge-intensive setting, we have adapted existing QFS datasets to create a new dataset, \dataname{}. For dataset collection, we first take query-summary pairs from the original QFS datasets. 
Instead of only summarizing manually-collected relevant documents to answer the query, we explore generating such a summary based on a much larger knowledge corpus that better fits realistic scenarios. In this work, we consider three types of knowledge corpora. The first is an internal collection, which is formed by combining all documents in the QFS datasets. However, the size of this collection is relatively limited as the size of the QFS datasets is small, which makes the new dataset less realistic. We then use commonly-used Wikipedia as an external large-scale collection. As no explicit evidence is included in this external collection, we combine both collections to form an augmented collection. 
As retrieval is necessary for the new setting,  we further provide relevance annotations to enable retrieval evaluation by recruiting crowdsourcing workers to annotate relevance labels.

To benchmark the proposed dataset \dataname{}, we consider two families of models: QFS models \cite{xu2020querysum,xu2020margesum} and retrieval-enhanced models \cite{izacard2021leveraging,lewis2020retrieval}. The performance of the former would indicate new challenges brought by the new setting compared with the original task. The latter have achieved state-of-the-art performance across many knowledge-intensive tasks such as open-domain question answering (QA) \cite{chen2017reading,kwiatkowski2019natural} and fact verification \cite{thorne2018fever}. Thus, their performance would indicate the efficacy of the existing methods in the new setting.
The experimental results on \dataname{} demonstrate that the summarization performance of QFS models is much lower than their performance on the original QFS task, showing that the new setting is much more challenging than the original QFS task and offers substantial room for improvement. Besides, we show that the retrieval-enhanced model outperforms QFS methods, indicating their better efficacy in realistic scenarios where knowledge comes from large-scale collections.

The contributions of this work are summarized as follows: (i) To our knowledge, we make the very first attempts to tackle QFS as a knowledge-intensive task, where relevant documents should be retrieved from a large-scale knowledge corpus. (ii) We construct a corresponding dataset \dataname{} to study the new setting, and relevance annotations are provided to enable retrieval evaluation as well. (iii) We benchmark the dataset with a variety of models, and the experimental results indicate the new challenges brought by the knowledge-intensive setting.

\section{Related Work}

\textbf{Query-focused summarization}. Due to the low resource setting, the previous work on QFS relied mainly on unsupervised approaches to extract salient sentences to produce the output summary~\cite{he2012document,li2017salience,liu2015multi}. 
With the increasing popularity of weak or distant supervision, recent studies~\cite{laskar2020wsl,xu2020querysum,xu2020margesum} have explored the adaption of neural models trained on other domains where large-scale resources are available.
For instance, \citet{laskar2020wsl} incorporated an answer selection model trained on a large-scale QA dataset, MS MARCO \cite{nguyen2016msmarco}. \citet{xu2020margesum} adopt a summarization model trained on a large-scale generic summarization dataset, CNN/DailyMail~\cite{hermann2015teaching}.
Another line of research aims to alleviate data scarcity by creating larger QFS datasets \cite{nema2012debatepedia,kulkarni2020aquamuse}. For instance, \citet{kulkarni2020aquamuse} build QFS datasets without any manual annotation using unsupervised semantic matching techniques. However, the existing datasets either focus on short summaries (i.e. one sentence) or are adapted from the datasets in other domains (i.e., QA). 
In contrast to all previous work, our work does not rely on any large resources from other domains. Instead, we tackle the QFS task in a more realistic knowledge-intensive setting.

\textbf{Knowledge-intensive tasks}. Knowledge-intensive tasks, including open-domain QA \cite{chen2017reading,kwiatkowski2019natural}, dialogue \cite{dinan2019wow}, and fact-checking \cite{thorne2018fever}, 
involve answering a given question from a large-scale knowledge base \cite{petroni2021kilt}. 
The closest KI task is called long-form question answering (LFQA) \cite{fan2019eli5,stelmakh2022asqa}, where outputs are informative answers to given questions. However, existing LFQA tasks either suffer from poor answer grounding \cite{krishna2021hurdle} or are restricted to ambiguous questions only \cite{min2020ambigqa}. Compared with them, our study has two main advantages: 1) expected answers are multi-document summaries, which are highly grounded in the documents; 2) queries reflect multiple information needs, and thus summaries are naturally longer in order to sufficiently cover various information.

\textbf{Open-domain multi-document summarization}. More recently, \citet{giorgi2022openmds} extends multi-document summarization (MDS) into an open-domain setting. 
This work conducts a similar investigation on MDS datasets \cite{fabbri2019multi,ghalandari2020wcep}. Compared with it, our study has two significant differences: 1) it uses gold summaries as pseudo-queries as queries are unavailable in MDS datasets, while we use high-quality queries written by human experts from QFS datasets; 2) It only considers internal documents within individual dataset, which restricts their analysis on a specific domain (e.g., news). While we consider both internal documents and large-scale external knowledge corpus, which makes our analysis better fit realistic scenarios.

\section{Task Formulation}
\label{sec:task}
We briefly introduce the original QFS task and then describe our proposed knowledge-intensive setting.

Let the tuple $(q, \mathcal{D}, s)$ denote a sample in the QFS dataset (a \textit{cluster}), where $q$ is a given query, $\mathcal{D}=\{d_1, \dotsc, d_M\}$ is a set of relevant documents, and $s$ is the gold summary. The goal of QFS is to produce the summary given the documents and the query as input: $(q, \mathcal{D}) \rightarrow s$. Note that $\mathcal{D}$ is collected manually by human annotators and usually includes tens of documents. 

In this work, we reformulate the QFS task to a knowledge-intensive task as $q \rightarrow s$. As shown in Figure \ref{fig:comparison}, we do not rely on the relevant document set $\mathcal{D}$. Instead, we assume that a large-scale knowledge corpus can be accessed: $\mathcal{K}=\{k_1, \dotsc\, k_N\}$. Normally, $\mathcal{K}$ consists of millions of documents, i.e., $N \gg M$.
Note that our goal remains to generate the summary $s$ with respect to the query $q$. However, different from the original setup, there is no evidence that the given documents are relevant to the query. More importantly, there are millions of candidate documents so it is impossible to consider all of them. Thus, an effective information retrieval model is necessary to supplement the extended setting, which is described in Section \ref{subsec:models}.

\section{\dataname{} Dataset}

In this section, we describe dataset collection and relevance annotation process. The dataset collection includes a collection of query-summary pairs and three knowledge corpora. The purpose of relevance annotation is to enable retrieval evaluation.

\subsection{Dataset Collection}

\textbf{Collecting query-summary pairs}. We build the dataset on the top of query-summary pairs $(q, s)$ on existing QFS resources. 
Specifically, we adopt DUC 2005-07 \cite{dang2006duc}, three standard QFS benchmark datasets. 
\footnote{Note that we take DUC as an example, but nothing prohibits exploring other QFS resources.}
The DUC datasets consist of three subsets collected for the Document Understanding Conferences (DUC) from 2005 to 2007. Each subset contains 45-50 clusters, where each cluster contains a query, a set of topic-related documents, and multiple reference summaries. As we assume that relevant documents are inaccessible, we only collect query-summary pairs. The documents will be used to build a knowledge corpus, which is described in the next paragraph.
For data split, we follow previous work~\cite{li2019document,laskar2020wsl} to use the pairs of the first two years as the training set and randomly select 10\% from the training set as the validation set. We leave the subset of 2007 as the test set. Finally, the training set, validation set, and test set contain 90, 10, and 45 examples, respectively.

\textbf{Building knowledge corpora}. We explore three alternatives of knowledge corpora for the query-summary pairs above. Specifically, we follow the standard data processing for large-scale knowledge sources \cite{karpukhin2020dense} to collect documents, where we split each origin document into non-overlapping context documents with 100 words maximum.
\footnote{Some studies also use passages to name the processed text, but we keep using documents to maintain the coherence of the paper.}
The first knowledge corpus, \corpusint{}, is an internal collection, where we take all clusters of DUC documents from the three years to form this collection, which results in about 32K documents in total. As the queries have an explicit connection with the documents, this collection can be considered an in-domain knowledge corpus, where relevant documents are relatively easy to find. 
However, as our main goal is to explore  knowledge-intensive QFS on a large-scale knowledge base, we consider the second external knowledge corpus named \corpusext{}. We use the Wikipedia dump in the KILT benchmark \cite{petroni2021kilt} to form this corpus, resulting in about 21 million documents in total. %
However, as \corpusext{} is a Wiki-based corpus, there is no guarantee that the collection contains sufficient evidence to answer the query since the reference summary is derived from the content of the original DUC documents.  %
To this end, we build an augmented knowledge corpus called \corpusaug{} by combining the previous two collections. We merge all the documents of \corpusint{} and \corpusext{} into this collection ensuring that summaries can be grounded in the collection and that the collection is large-scale to make the task challenging.

\subsection{Relevance Annotation}
\label{subsec:retrilabel}

As the retrieval is performed for our knowledge-intensive setting, it is important to evaluate the performance so we can better analyze its effect on summarization.
However, the DUC datasets do not come with relevance labels to designate which document provides evidence towards the final summary. 
A possible solution to that would be to do a lexical match between the individual summary sentences and the available document collection, however, such an automatic evaluation has its disadvantages (e.g. a summary sentence may appear in a document but outside the right context, lexical overlap may pay attention to insignificant words in the summary, and there may be a lexical gap between semantically similar sentences). 
Therefore, we collect human annotations through Amazon Mechanical Turk (MTurk) to create our evaluation set. Following the TREC document retrieval task \cite{craswellC2020trecdeep}, we apply retrieval models to choose top-ranked documents that are used fro annotation. The MTurk workers are asked to label a relevance score for each query-document pair using a 4-point scale. More details on the annotation process are provided in Appendix \ref{sec:app_annotation}.

\begin{table}[t]
\centering
\setlength\tabcolsep{6pt}
\caption{Comparisons of retrieval models on the test set of \dataname{} with three knowledge corpora. * means that the improvement is statistically significant (a two-paired t-test with p-value < 0.01).}
\label{tab:retrieval}
\begin{tabularx}{0.45\textwidth}{l *{5}{l}}
\toprule[1pt]
Corpus & Model  & P@10 & P@50 & R@10 & R@50 \\
\hline
\multirow{2}{*}{\corpusint{}} & BM25 & \textbf{16.00}$^*$  & \textbf{11.96} & \textbf{8.24}$^*$ & \textbf{30.19}$^*$ \\
& DPR & 11.56 & 11.51 & 5.97 & 27.80 \\
\hline
\multirow{2}{*}{\corpusext{}} & BM25 & 12.67  & 8.58 & \textbf{7.00}  & 22.22 \\
& DPR & \textbf{13.78}$^*$ & \textbf{12.58}$^*$ & 6.95  & \textbf{29.69}$^*$ \\
\hline
\multirow{2}{*}{\corpusaug{}} & BM25 & 12.22 & 8.98 & 6.84 & 23.21 \\
& DPR & \textbf{14.44}$^*$ & \textbf{12.53}$^*$ & \textbf{7.41}$^*$ & \textbf{30.32}$^*$ \\
\bottomrule[1pt]
\end{tabularx}
\end{table}

\begin{table}[b]
\centering
\caption{Comparisons of QFS models on the original document sets (\textsc{Origin}) and our knowledge corpus (\corpusint{}). The results inside the bracket denote the percentage decrease ($\downarrow$) when the models are adapted to the knowledge-intensive setting.}
\label{tab:qfs_duc}
\begin{tabularx}{0.49\textwidth}{l *{4}{c}}
\toprule[1pt]
Corpus & Model  & ROUGE-1 & ROUGE-2 & ROUGE-SU4 \\
\hline
\multirow{2}{*}{\textsc{Origin}} & \querysum{} & 43.25 & 11.58 & 16.76  \\
& \margesum{} & 42.52 & 11.77 & 16.73   \\
\hline
\multirow{2}{*}{\corpusint{}} & \querysum{}  & 36.08 (17\%) & 7.54 (35\%) & 12.69 (24\%) \\
& \margesum{} & 37.99 (11\%) & 9.13 (22\%) & 14.29 (15\%)  \\
\bottomrule[1pt]
\end{tabularx}
\end{table}
\begin{table*}[t]
    \small
    \centering
    \caption{Comparison of summarization performance on the three knowledge corpora. R-1, R-2 and R-SU4 stand for ROUGE-1, ROUGE-2 and ROUGE-SU4, repectively. Note that the results of BARTScore are normalized with the exponential function. * indicates the improvement to the best QFS model is statistically significant (a two-paired t-test with p-value < 0.01).}
    \label{tab:main_duc}
    \begin{adjustbox}{max width=\textwidth}
    {
    \begin{tabular}{l *{15}{l}}
        \toprule
        & \multicolumn{5}{c}{\corpusint{}} & 
        \multicolumn{5}{c}{\corpusext{}} &
        \multicolumn{5}{c}{\corpusaug{}} \\
        \cmidrule(lr){2-6} 
        \cmidrule(lr){7-11}
        \cmidrule(l){12-16}
        \textbf{Model}   & R-1 & R-2 & R-SU4 & BERT$_{\text{score}}$ & BART$_{\text{score}}$ & R-1 & R-2 & R-SU4 & BERT$_{\text{score}}$ & BART$_{\text{score}}$ & R-1 & R-2 & R-SU4 & BERT$_{\text{score}}$ & BART$_{\text{score}}$  \\ 
        \midrule
        Oracle+FiD & 44.14 & 12.63 & 17.50 & 24.35 & 3.77 & 38.79 & 8.38 & 14.27 & 15.77 & 3.25 & 44.08 & 12.58 & 17.42 & 24.98 & 3.77 \\
        \cmidrule(lr){1-16} 
       \querysum{}  & 36.08 & 7.54 & 12.69 & 8.53 & 3.08 & 31.07 & 4.52 & 10.19 & 2.03 & 2.72 & 32.64 & 5.47 & 11.09 & 2.96 & 2.84  \\
        \margesum{} & 37.99 & 9.13 & 14.29 & 11.49 & 3.09 & 34.37 & 6.52 & 12.20 & 5.86 & 2.84 & 36.69 & 8.07 & 13.52 & 8.69 & 3.01 \\
        \cmidrule(lr){1-16} 
        RAG & 28.92 & 5.72 & 10.07 & 12.60 & 3.19 & 32.28 & 5.21 & 10.76 & 7.99 & 3.03 & 27.06 & 4.57 & 9.00 & 8.28  & 3.03  \\
        BM25+FiD & \textbf{42.44}$^*$ & \textbf{11.25}$^*$ & \textbf{16.54}$^*$ & \textbf{21.42}$^*$ & \textbf{3.61} & \textbf{38.80}$^*$ & \textbf{8.36}$^*$ & \textbf{14.22}$^*$ & \textbf{15.69}$^*$ & \textbf{3.25} & \textbf{41.40}$^*$ & \textbf{10.83}$^*$ & \textbf{16.07}$^*$ & \textbf{19.99}$^*$  & \textbf{3.47} \\
        DPR+FiD & 41.46 & 10.72 & 15.91 & 21.41 & 3.60 & 38.56 & 8.01 & 14.10 & 15.50 & 3.22 & 39.99 & 9.42 & 15.06 & 18.04 & 3.38 \\
        \bottomrule
    \end{tabular}
    }
    \end{adjustbox}
\end{table*}

\section{Experiments}
\label{sec:exp}

\subsection{Models}
\label{subsec:models}

\textbf{QFS models}. To illustrate new challenges brought by the proposed knowledge-intensive setting, we first evaluate the models designed for original QFS tasks: (i) \querysum{} \cite{xu2020querysum} is an extractive QFS model leveraging distant supervision signals from trained QA models to select salient sentences as the summary. (ii) \textsc{MargeSum} \cite{xu2020margesum} is an abstractive QFS model that employs generative models trained on generic summarization resources to boost sentence ranking and summary generation, achieving competitive performance in the weakly supervised setting. Implementation details about QFS models are provided in Appendix \ref{subsec:app_qfsmodels}

\textbf{Retrieval-enhanced models}. We also adopt retrieval-enhanced generation models in the new setting: 
(i) Retrieval-Augmented Generation (RAG) \cite{lewis2020retrieval} is a generative model in which document retrieval and summary generation are learned jointly. As RAG models have two variants including RAG-Token and RAG-Sequence, we mainly report the results of RAG-Sequence as it performs better in open-domain QA. 
(ii) Fusion-in-Decoder (FiD) \cite{izacard2021leveraging} is an improved encoder-decoder architecture which has achieved state-of-art performance in some knowledge-intensive tasks \cite{asai2022evidence}. In this model, encoded representations for top-$k$ retrieved documents are first concatenated and then fed into the decoder to generate the output answer. We explore the effect of two commonly-used retrieval models: BM25 \cite{robertson2009bm25} and Dense Passage Retrieval (DPR) \cite{karpukhin2020dense}. 
More implementation details about the models are provided in Appendix \ref{subsec:app_remodels}

\subsection{Evaluation Metrics}

As our knowledge-intensive setting includes a retrieval module and a summarization module, we evaluate both modules using different metrics. For retrieval evaluation, we use precision@$k$ (P@$k$) and recall@$k$ (R@$k$) where $k$ denotes the cut-off. For summarization evaluation, we use standard ROUGE-1, ROUGE-2, and ROUGE-SU4 ~\cite{lin2004rouge}.
We also report semantic metrics including BERTScore \cite{zhang2020bertscore} and BARTScore \cite{yuan2021bartscore}.
More details on the metrics are given in Appendix \ref{sec:app_metric}.

\subsection{Results and Analysis}
\label{subsec:results}

Our experiments mainly are designed to answer the following research questions (RQs): (i) \textbf{RQ1:} What is the performance of retrieval models in the knowledge-intensive setting? (ii) \textbf{RQ2:} Is this setting much more challenging than the original QFS? (iii) \textbf{RQ3:} How do QFS models and retrieved-enhanced models perform in the \dataname{} dataset?

\subsubsection{Retrieval Results}

To answer \textbf{RQ1}, we compare the retrieval performance of BM25 and DPR on the test set of \dataname{} (see Appendix \ref{sec:app_annotation} for details of relevance annotation). 
As shown in Table \ref{tab:retrieval}, BM25 outperforms DPR on \corpusint{} than \corpusaug{}. One possible explanation is that relevant documents in internal knowledge corpus \corpusint{} contain more relevant keywords. These keywords can be easier retrieved by BM25 which is a term-based IR method.
While DPR performs better on \corpusext{}, Since it is pre-trained on the dump of Wikipedia so semantic representations of documents are learned better, which leads to retrieval improvement.
The fairly low precision and recall scores across all three collections show that knowledge retrieval is challenging, and more effort is required to boot the performance.

\subsubsection{Summarization Results}

To answer \textbf{RQ2},
we compare the start-of-the-art QFS models (distantly supervised) between original document sets (\textsc{Origin}) and our \corpusint{}. Note that the only difference between them is that \textsc{Origin} only contains a small set of relevant documents while \corpusint{} consists of a large number of candidate documents in the datasets. As shown in Table \ref{tab:qfs_duc}, we find that the performance of QFS models drops largely when they are performed in the knowledge-intensive setting. This is because QFS systems are developed for closed-domain retrieval and suffer from the distributional shift when they perform knowledge retrieval in an open-domain setting. This shows the proposed dataset is challenging, and more research effort is needed to adapt close-domain QFS models to our setting.

To answer \textbf{RQ3}, we benchmark our dataset with QFS models and retrieval-enhanced models. As shown in Table \ref{tab:main_duc},
We first report the results of an Oracle+FiD model which serves as an upper-bound performance.
To obtain Oracle documents, we first compute ROUGE scores between all documents in the corresponding knowledge corpus and gold summary, and then we select top-$k$ documents as Oracle documents based on their scores. 
We observe a significant increase in the performance of the Oracle+FID especially on the \corpusaug{}. This shows that the retrieval performance in large-scale collections really
affects the summarization performance.
We find that FiD-based models outperform QFS models, showing benefits from fine-tuning on the datasets.
We notice that the performance of the RAG-based model shows inferior performance, potentially due to the small size of the training data.

\section{Conclusion}
\label{conclu}
In this paper, we address the QFS task in a knowledge-intensive setting, where relevant documents are not readily available and should be retrieved from a large-scale knowledge corpus. To this end, we build a new dataset based on the existing DUC datasets. In this dataset, we construct three knowledge corpora to enable document retrieval. To evaluate retrieval performance, we further provide relevance annotations.
We benchmark the dataset with QFS models and retrieval-enhanced models, the experimental results show that the knowledge-intensive setting is much more challenging than the original setting.
In future work, we consider generating relevance labels using some domain adaption methods, in which human annotation may not be required. %

\balance

\bibliographystyle{ACM-Reference-Format}
\bibliography{main}

\appendix

\section{Details on Annotation Process}
\label{sec:app_annotation}

As the main goal of the annotation process is to identify relevant documents that support the summary, we begin with filtering out irrelevant documents, which are the majority of millions of candidate documents in the knowledge corpus.
Following the TREC document retrieval task \cite{craswellC2020trecdeep}, we apply retrieval models to choose top-ranked documents.
In particular, we consider two different but effective retrieval approaches, BM25 \cite{robertson2009bm25} and Dense Passage Retrieval (DPR) \cite{karpukhin2020dense}. 
We first perform both BM25 and DPR on \corpusint{}, \corpusext{} and \corpusaug{}, and construct top-50 pools, which result in a maximum of 300 candidate documents for each query. Not that duplicate documents may exist in the pool and would be removed. As there are 45 queries in our test set, we end up with 7761 query-document pairs to be annotated. The MTurk workers are asked to label a relevance score for each pair using a 4-point scale. 
We design a web annotation interface shown to MTurk workers, which contains a query, a reference summary, and at most five documents. We ask workers to label whether a document contains either part of the summary or evidence to a query, adapting the annotation instructions in \citet{craswellC2020trecdeep}. 
The judgments are on a 4-point scale from 0 to 3:
\begin{itemize}
    \item \textbf{Irrelevant (0)}: the document has nothing to do with the query.
    \item \textbf{Weakly related (1)}: the document seems related to the query but fails to contain any evidence in the summary.
    \item \textbf{Related (2)}: the document provides unclear information related to the query, but human inference may be needed.
    \item \textbf{Relevant (3)}: the document explicitly contains evidence that is part of the summary.
\end{itemize}
Prior to the collection of the annotations we performed two pilot studies to improve our annotation interface and the instructions. We collect three annotations per document and use a majority vote to define the label. The distribution of relevance labels is 837 relevant, 5811 related, 1097 weakly related, and 16 irrelevant documents. 
We also measure inter-annotator agreement (IAA) using Fleiss Kappa \cite{fleiss1971measuring}. A Fleiss Kappa score of 0.245 was obtained, which indicates a fair agreement. 

From the annotation results, we find that although most retrieved documents are related to the queries to some extent,
there are only 837 documents ($10.8\%$) that contained explicit fragments of the summary. 
To validate whether the documents with relevance labels equal to 3 are sufficient to generate the summaries, we compared the summarization performance when documents annotated as perfectly relevant were provided as input and the performance when documents annotated either as perfectly relevant or relevant, with the latter offering no extra performance gains over the former.

\section{Implementation Details}
\label{sec:app_models}

\subsection{QFS Models}
\label{subsec:app_qfsmodels}

As QFS models have achieved competitive performance in the original DUC datasets, we directly use them to generate summaries without any further training. For inference, as they include a retrieval module that performs in-domain evidence estimation, we can directly implement the models on the collection \corpusint{}. However, their retrieval models are not designed to handle large-scale, open-domain knowledge bases, such as \corpusext{}. Therefore, we first retrieve top-1000 documents using BM25 for each query and then feed them as inputs to their customized retrieval models. We follow the original papers to set the hyper-parameters. All models are conducted on a single Titan-XP GPU.

\subsection{Retrieval-Enhanced Models}
\label{subsec:app_remodels}

As retrieval-enhanced generation models are designed for open-domain QA where answers are much shorter. For fair comparisons, we fine-tune them in our training set. To be specific, there are 4 gold summaries in the datasets. we randomly choose a summary to construct the input-output pairs to construct each training batch with a batch size of 1.
For the hyper-parameters of FiD, we set $k$ for top-$k$ retrieved documents to 50 after a small validation study. Following the origin paper, we use $\mathrm{T5_{base}}$ \cite{raffel2020exploring} as the base model. For DPR, we use the version trained on the KILT benchmark \cite{petroni2021kilt}. As for RAG, we follow the original paper to set the number of retrieved documents to 10 \cite{lewis2020retrieval}. For fine-tuning both models, we use Adam \cite{kingma2014adam} with a learning rate of $10^{-4}$ and the maximum number of training epochs to 30. We also apply early-stopping strategies to avoid over-fitting when we find there is no improvement in the validation set within 10 training steps. For inference, we set the maximum decoded length and beam size to 250 and 5, respectively. All experiments are conducted on a single Titan-XP GPU.

\section{Details on Evaluation Metrics}
\label{sec:app_metric}

\subsection{Retrieval Evaluation}
\label{subsec:app_retrieval}

To evaluate retrieval effectiveness, i.e., the ability of the QA systems to identify and retrieve evidence to be used to compose the final answer,
we use standard information retrieval (IR) metrics, and in particular precision@$k$ (P@$k$) and recall@$k$ (R@$k$) where $k$ denotes the cut-off.
For P@$k$ and R@$k$, we mapped the judgment level 3 to relevant and judgment levels 0-2 to irrelevant and reported the corresponding results for $k \in \{10, 50\}$.

\subsection{Summarization Evaluation}
\label{subsec:app_summeval}

In our experiments, we measure the accuracy of generated summaries against gold summaries.
Following previous work in the QFS task \cite{xu2020querysum,laskar2020wsl}, we employ the standard ROUGE-1, ROUGE-2, and ROUGE-SU4 ~\cite{lin2004rouge}. They evaluate the accuracy in terms of uni-grams, bi-grams, and bi-grams with a maximum skip distance of 4, respectively.  
We report the $\mathrm{F_1}$ score with a maximum length of 250 words for the metrics. 
Note that there are multiple reference summaries in DUC datasets, we take the average of the ROUGE scores for all summaries as the final reported ROUGE score.

Although ROUGE-based metrics are commonly used, they fail to capture semantic similarities since they only measure lexical overlap. Recent semantic metrics, such as BERTScore \cite{zhang2020bertscore} and BARTScore \cite{yuan2021bartscore} have been shown to be strongly correlated with human judgments. Therefore, we also report both metrics in our experiments. Specifically, we report the F1 score of BERTscore and use recommended \texttt{deberta-xlarge-mnli} \cite{he2021deberta} as the backbone of BERTScore. For BARTScore, we use the recall score of BARTScore as it is more suitable for summarization tasks \cite{yuan2021bartscore}, where we use \texttt{bart-large-cnn} as the base model. As the outputs of BARTScore are negative log probabilities that are difficult to explain, we normalize them with the exponential function.

\end{document}